\documentclass{article}

\usepackage{microtype}
\usepackage{graphicx}
\usepackage{subfigure}
\usepackage{float}
\usepackage{booktabs} 

\usepackage{hyperref}

\usepackage[accepted]{icml2025}

\usepackage{colortbl}

\usepackage{amsmath}
\usepackage{amssymb}
\usepackage{mathtools}
\usepackage{amsthm}
\usepackage{bm}
\usepackage{enumitem}
\usepackage{multirow}
\usepackage[capitalize,noabbrev]{cleveref}
\usepackage{balance}

\theoremstyle{plain}

\theoremstyle{definition}

\theoremstyle{remark}

\definecolor{red}{RGB}{192,0,0}
\definecolor{blue}{RGB}{48, 85, 151}
\newcommand{\addvalue}[3]{\hspace{#3mm}#1\scalebox{0.8}{\textcolor{blue}{{\textbf{+#2}}}}}
\newcommand{\minusvalue}[3]{\hspace{#3mm}#1\scalebox{0.8}{\textcolor{red}{{\textbf{-#2}}}}}
\newcommand{\conf}[1]{\scalebox{0.7}{\textcolor{gray}{({#1})}}}
\usepackage{ulem}

\usepackage[textsize=tiny]{todonotes}

\renewcommand{\algorithmiccomment}[1]{\hfill $\triangleright$ #1}

\icmltitlerunning{DAMA: Data- and Model-aware Alignment of Multi-modal LLMs}

\begin{document}

\twocolumn[
\icmltitle{DAMA: Data- and Model-aware Alignment of Multi-modal LLMs }

\icmlsetsymbol{equal}{*}

\begin{icmlauthorlist}
\icmlauthor{Jinda Lu}{ustc}
\icmlauthor{Junkang Wu}{ustc}
\icmlauthor{Jinghan Li}{ustc}
\icmlauthor{Xiaojun Jia}{ntu}
\icmlauthor{Shuo Wang}{ustc}
\icmlauthor{Yifan Zhang}{iac}
\icmlauthor{Junfeng Fang}{nus}
\icmlauthor{Xiang Wang}{ustc}
\icmlauthor{Xiangnan He}{ustc}
\end{icmlauthorlist}

\icmlaffiliation{ustc}{University of Science and Technology of China}
\icmlaffiliation{ntu}{Nanyang Technological University}
\icmlaffiliation{nus}{National University of Singapore}
\icmlaffiliation{iac}{Institute of Automation, Chinese Academy of Sciences}

\icmlcorrespondingauthor{Xiang Wang}{xiangwang1223@gmail.com}

\vskip 0.3in
]
\printAffiliationsAndNotice{}  

\begin{abstract}
Direct Preference Optimization (DPO) has shown effectiveness in aligning multi-modal large language models (MLLM) with human preferences. 
However, existing methods exhibit an imbalanced responsiveness to the data of varying hardness, tending to overfit on the \textit{easy-to-distinguish} data while underfitting on the \textit{hard-to-distinguish} data. 
In this paper, we propose \textbf{Da}ta- and \textbf{M}odel-\textbf{a}ware DPO (DAMA) to dynamically adjust the optimization process from two key aspects: (1) a data-aware strategy that incorporates data hardness, and (2) a model-aware strategy that integrates real-time model responses. By combining the two strategies, DAMA enables the model to effectively adapt to data with varying levels of hardness.
Extensive experiments on five benchmarks demonstrate that DAMA not only significantly enhances the trustworthiness, but also improves the effectiveness over general tasks. For instance, on the Object HalBench, our DAMA-7B reduces response-level and mentioned-level hallucination by 90.0\% and 95.3\%, respectively, surpassing the performance of GPT-4V. Code is available at: \url{https://github.com/injadlu/DAMA}

\end{abstract}
\section{Introduction}
\label{sec:intro}

\begin{figure}[t]
\vskip 0.2in
\begin{center}
\centerline{\includegraphics[width=\columnwidth]{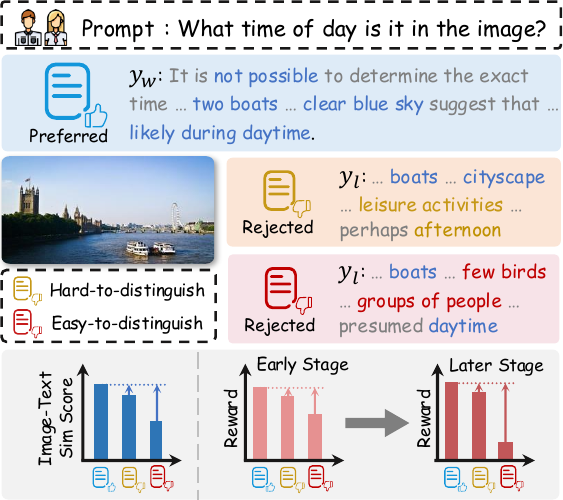}}
\caption{ (1) Preference data (Prompt, Image, Preferred response $y_w$, Rejected response $y_l$) with different hardness: ``easy-to-distinguish'' data denotes a large Image-Text sim score gap between $y_l$ and $y_w$; ``hard-to-distinguish'' data indicates a low score gap between $y_l$ and $y_w$.
(2) Implicit reward across the optimization stage: the reward gap for ``easy-to-distinguish" data enhances significantly during optimization, while for ``hard-to-distinguish" data, the gap remains low.}
\label{fig:figure-1}
\end{center}
\vskip -0.2in
\end{figure}

Recent advances in Multimodal Large Language Models (MLLMs) have demonstrated remarkable visual understanding capabilities on the basis of large language models \cite{LLaVA, InternVL, QwenVL}. 
However, despite their effectiveness, the hallucination issue --- generating outputs inconsistent with the image content and human preference --- limits their broader applicability \cite{VCD, LRV-Instruction, RLHF-V}. 
To address this, direct preference optimization (DPO) \cite{DPO} has been adapted into MLLM alignment \cite{RLAIF-V, mDPO, CLIP-DPO}, achieving encouraging performance with moderate computational costs.

DPO methods \cite{LLaVA-RLHF, RLAIF-V} collect preference data consisting of an image, a prompt, and two responses ($y_w, y_l$). The preferred response $y_w$ is better aligned with the visual content (large Image-Text similarity score), while the rejected response $y_l$ contains more hallucinated content (small Image-Text similarity score). 
They prioritize preferred responses $y_w$ over the rejected ones $y_l$ using DPO with a hyperparameter $\beta$, which balances retaining the reference model $\pi_\text{ref}$ and incorporating new preferences into the updated model $\pi_{\bm{\theta}}$ \cite{TPO, V-dpo}.

However, our analysis reveals that current methods exhibit imbalanced responsiveness in handling the data with varying hardness during the optimization process, resulting in suboptimal performance.
As illustrated in Figure \ref{fig:figure-1}, for ``easy-to-distinguish" data (large Image-Text similarity score gap between preferred $y_w$ and rejected $y_l$), the reward gap amplifies during training, indicating stronger alignment.
Conversely, for ``hard-to-distinguish'' data (small Image-Text similarity score gap), the reward gap stagnates, suggesting limited capability to distinguish $y_w$ from $y_l$. 
This implies that current methods, which employ optimization strategies using a static $\beta$ across data with varying hardness \cite{LLaVA-RLHF, RLAIF-V}, could fail to capture the learning dynamics inherent in multimodal preference data.

To address this imbalanced responsiveness issue, we propose \textbf{Da}ta- and \textbf{M}odel-\textbf{a}ware direct preference optimization (\textbf{DAMA}), which dynamically adapts $\beta$ to both data hardness and model's responsiveness, enabling adaptively adjust model's learning behavior based on the inherent preference dynamics and model's real-time responses. 
Specifically, we propose two novel mechanisms:

\textbf{Data-aware Preference Optimization. (Section \ref{subsec:Data-anchored Preference Optimization})}: 
    We quantify data hardness via CLIP-based image-text similarity scores \cite{CLIP}, decomposing responses into sub-sentences for granularity-aware estimation.
    Then we normalize and transform the scores into probabilities to enable effective hardness estimation.
    By dynamically scaling $\beta$ inversely with hardness, we enforce stronger regularization on ``easy-to-distinguish'' samples (large $\beta$) while relaxing constraints for ``hard-to-distinguish'' ones (small $\beta$), preventing overfitting and underfitting, respectively.

\textbf{Model-aware Preference Optimization. (Section \ref{subsec:Model-aware Preference Optimization})}: 
    We estimate the model's responsiveness through the reward gaps between the preferred and rejected responses. Moreover, to improve the robustness of the estimation, we normalize the reward gaps and filter out outliers. 
    Similarly to Section \ref{subsec:Data-anchored Preference Optimization}, we incorporate the estimated responsiveness into the optimization process by dynamically modifying $\beta$ according to it, enabling the model to effectively optimize according to its current responsiveness.

By combining these strategies via element-wise multiplication, DAMA enables real-time adaptation to both data hardness and model responsiveness, demonstrating strong alignment performance across various evaluation benchmarks.
Our contributions are summarized as follows:
\begin{itemize}[leftmargin=*]
    \item We introduce a data-aware preference optimization strategy to integrate the data hardness into the preference optimization process, enabling the model to adaptively optimize based on the data hardness.
    \item We propose a model-aware preference optimization strategy to incorporate the real-time model response into the preference optimization process, facilitating the model to effectively optimize based on its current responsiveness.
    \item By combining the data- and model-aware strategies, we adaptively refine the optimization process, achieving significant alignment performance, as demonstrated by extensive empirical evaluations.
\end{itemize}

\begin{figure*}[t]
\vskip 0.2in
\begin{center}
\centerline{\includegraphics[width=\textwidth]{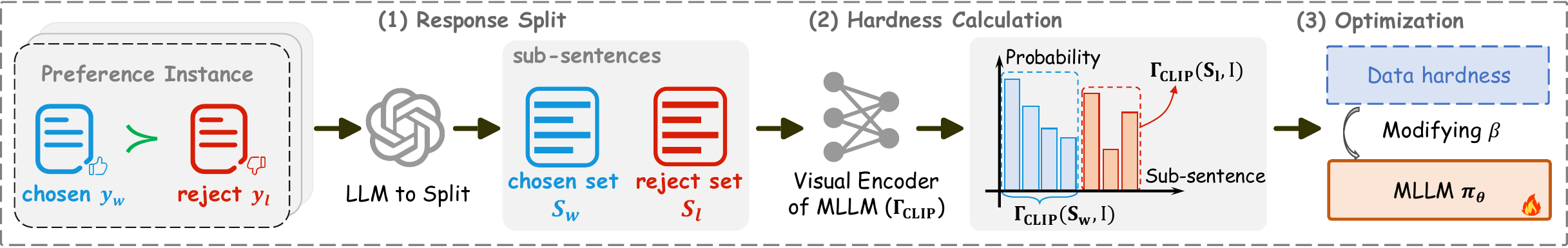}}
\caption{Overview of our data-aware preference optimization. For each preference instance: (1) We first break the preferred and rejected response into sub-sentences by prompting a large language model (LLM); 
(2) Next, we estimate the similarity scores between each sub-sentence and the given image using the CLIP classifier, and then calculate the differences between the preferred and rejected response as the hardness of the data; 
(3) Finally, we incorporate the estimated hardness into the preference optimization process by modifying $\beta$ in Equ~\eqref{equ:dpo}, allowing the model to adjust based on the data hardness.}
\label{fig:pipleine-vlm}
\end{center}
\vskip -0.2in
\end{figure*}

\section{Preliminary}
\label{sec:preliminary}
In this section, we briefly review the MLLM preference learning procedure, which starts by sampling pairwise preference data with a supervised fine-turned (SFT) model, and then optimizes on such preference data. Specifically, we categorize this process into the following aspects:

\noindent \textbf{Supervised Fine-Tuning (SFT).}
Preference learning of an MLLM $\bm{\pi}$ begins with an SFT model $\bm{\pi}_{\text{SFT}}$. Concretely, the SFT process fine-tunes the pre-trained MLLM model with millions of multi-modal question-answer pairs to align LLM with multi-modal tasks. 
After this process, we construct preference data by sampling pair-wise preference responses from $\bm{\pi}_{\mathrm{SFT}}$, formalized as $(y_w, y_l) \sim \bm{\pi}_{\mathrm{SFT}}(y|x,\mathcal{I})$, where $(\mathcal{I}$ denotes the image and $x$ is the prompt question. 
Meanwhile, $(y_w, y_l)$ are labeled as preferred and less preferred responses by humans, formalized as $(y_w \succ  y_l | \mathcal{I}, x)$.

\noindent \textbf{RLHF with Reward Models.}
Given pair-wise preference data $(y_w, y_l) \sim \bm{\pi}_{\mathrm{SFT}}(y|x,\mathcal{I})$, the preference learning process can be described in 2 stages: reward modeling and preference optimization. 
Specifically, the reward model $r_{\bm{\theta}}(y|\mathcal{I}, x)$ is defined to rank the model responses by learning to distinguish $y_w$ from $y_l$, and the preference optimization aims to distill the preference knowledge into MLLM. 
To learn a reward model, pioneering work \cite{rlhf} employs the Bradley-Terry model \cite{BT_model} to model the pair-wise preference distribution as:
\begin{equation}
\resizebox{.9\hsize}{!}{
\begin{math}
\begin{aligned}
    \mathrm{P}(y_w \succ  y_l|\mathcal{I}, x) & =  \sigma(r^{*}(y_w|\mathcal{I}, x)- (r^{*}(y_l|\mathcal{I}, x)) \\
     & = \frac{\mathrm{exp}(r^{*}(y_w|\mathcal{I}, x))}{\mathrm{exp}(r^{*}(y_w|\mathcal{I}, x))+\mathrm{exp}(r^{*}(y_l|\mathcal{I}, x))}.
\end{aligned}
\end{math}
}
\end{equation}

Thus, the learning process can be achieved by minimizing the negative log-likelihood $-\mathrm{logP}(y_w \succ y_l|\mathcal{I}, x)$ over the preference data with the parametrized reward model $r_{\bm{\phi}}(y_w|\mathcal{I}, x)$ initialized as $\bm{\pi}_{\mathrm{SFT}}$ with a simple linear layer to produce reward prediction. 
With the well-optimized reward model $r_{\phi}^{*}(y|\mathcal{I}, x)$, prior work \cite{rlhf} proposes to employ policy optimization algorithms in RL such as PPO \cite{PPO} to maximize the learned reward with KL-penalty, which can be formalized as:
\begin{equation}
\label{equ:ppo}
\begin{aligned}
    \underset{\bm{\pi}_{\theta}}{\text{max}} & \  \mathbf{E}_{(\mathcal{I},x) \sim \mathcal{D}, y \sim \bm{\pi}_{\theta}(\cdot|\mathcal{I}, x)} [r_{\phi}^{*}(y|\mathcal{I}, x)] \\
    & -\beta \mathbb{D}_{\mathbf{KL}}[\bm{\pi}_{\theta}(y|\mathcal{I},x)||\bm{\pi}_{\text{ref}}(y|\mathcal{I},x)], 
\end{aligned}
\end{equation}
where the fixed reference model $\bm{\pi}_{\text{ref}}$ is parameterized as $\bm{\pi}_{\text{SFT}}$, and the hyper-parameter $\beta$ controls the deviation of $\bm{\pi}_{\theta}$ from $\bm{\pi}_{\text{ref}}$ during the optimization process.

\noindent \textbf{Direct Preference Optimization (DPO).}
To relieve the high computational complexity of reward training in RLHF, DPO \cite{DPO} is proposed, which provides a simple way to directly optimize $\bm{\pi}_{\theta}$ with the pair-wise preference data, without parametrized reward model. Specifically, the DPO loss can be described as:
\begin{equation}
\label{equ:dpo}
\begin{aligned}
    \mathcal{L}_{\mathrm{dpo}} = - \bm{\mathrm{E}}_{(\mathcal{I},x, y_{w}, y_{l})} [ {\log \sigma}( & \beta \log \frac{{\pi}_{\bm{\theta}}(y_{w}|\mathcal{I},x)}{{\pi}_{\mathrm{ref}}(y_{w}|\mathcal{I},x)} \\
    - & \beta \log \frac{{\pi}_{\bm{\theta}}(y_{l}|\mathcal{I},x)}{{\pi}_{\mathrm{ref}}(y_{l}|\mathcal{I},x)}) ].
\end{aligned}
\end{equation}
\section{Approach}
\label{sec:approach}
In this section, we describe our DAMA in detail. Specifically, we first illustrate our data-aware preference optimization, then we describe our model-aware preference optimization, and finally, we show our combination strategies for robust preference optimization. Our approach algorithm is listed in Algorithm. \ref{algorithm}.

\subsection{Data-aware Preference Optimization}
\label{subsec:Data-anchored Preference Optimization}
An overview of our data-aware preference optimization is shown in Figure~\ref{fig:pipleine-vlm}. Given a preference instance from the dataset $\mathcal{D}$ as $\{ (\mathcal{I}, x, y_{w}, y_{l})\} \sim \mathcal{D}$, where $\mathcal{I}$, $x$, $y_{w}$, $y_{l}$ denotes the image, question, preferred response, and rejected response, respectively, it firstly splits the responses into simple and self-contained sub-sentences. Next, it calculates the image-text similarity scores between the sub-sentences and the image by the CLIP classifier. Then, it combines the scores of each response and compares the difference between the preferred and rejected responses as the data hardness. This hardness is embedded into the preference optimization process by modifying the $\beta$ in Equ~\eqref{equ:dpo}.
The following are detailed descriptions.

We employ the CLIP classifier $\mathbf{\Gamma}_{\mathrm{CLIP}}$ \cite{CLIP}, to calculate similarity scores.
For each preference instance, we aim to effectively capture the similarity between the responses $(y_{w}, y_{l})$ with the given image $\mathcal{I}$, while alleviating the 77 token length constraints in CLIP. To achieve this, we decompose the complex responses, which contain various objects and relations, into simple and self-contained sub-sentences. 
Concretely, we prompt the open-source large language model, such as LLaMA-3 \cite{LLaMA3}, to split $(y_{w}, y_{l})$ into sub-sentences $\mathbf{S}_{w} = \{ \mathbf{S}_{w,j} | j = 1,2, \dots p\}$ and $\mathbf{S}_{l} = \{ \mathbf{S}_{l,k} | k = 1,2, \dots q\}$, where $p$ and $q$ denotes the number of sub-sentences for $\mathbf{S}_{w}$ and $\mathbf{S}_{l}$.

Subsequently, we employ the CLIP classifier $\mathbf{\Gamma}_{\mathrm{CLIP}}$ to calculate the similarity score between the given image $\mathcal{I}$ and the sub-sentences $\mathbf{S}_{w}, \mathbf{S}_{l}$ as:
\begin{equation}
\label{equ:subsentence-score}
\begin{aligned}
\mathbf{C}_{w} &= \left[ \mathbf{\Gamma}_{\mathrm{CLIP}}(\mathcal{I}, \mathbf{S}_{w,j}) \right]_{j=1}^{p}, \\
\mathbf{C}_{l} &= \left[ \mathbf{\Gamma}_{\mathrm{CLIP}}(\mathcal{I}, \mathbf{S}_{l,k}) \right]_{k=1}^{q},
\end{aligned}
\end{equation}
where $\mathbf{C}_{w} \in \mathbb{R}^{p}$ and $\mathbf{C}_{l} \in \mathbb{R}^{q}$ represents the corresponding similarity scores of $\mathbf{S}_{w}$ and $ \mathbf{S}_{l}$, respectively. 
To effectively quantify the difference between preferred response $\mathbf{C}_{w}$ and rejected response $\mathbf{C}_{l}$ for each instance, 
we normalize the corresponding score by the softmax probabilities as :
\begin{equation}
\label{equ:subsentence-prob}
\left[ \begin{array}{c}
\mathbf{P}_{w} \\
\mathbf{P}_{l}
\end{array} \right] = 
\mathbf{Softmax}   \left (  \left[ \begin{array}{c}
    \mathbf{C}_{w} \\
    \mathbf{C}_{l}
    \end{array} \right]  \right ) ,
\end{equation}
$\mathbf{P}_{w} \in \mathbb{R}^{p}$ and $\mathbf{P}_{l} \in \mathbb{R}^{q}$ represents the probabilites. 
The difference between the preferred and rejected probabilities demonstrates the data hardness. A large difference implies that the preference data is ``easy-to-distinguish'', where the rejected response includes more elements that are not present in the image, conversely, a small difference suggests that the preference data is ``hard-to-distinguish'', and the rejected response exhibits minimal hallucination. 
Then we define the hardness based on the probabilities difference as:
\begin{align}
    \label{equ:difference}
        \delta = & \sum_{j=1}^{p} \mathbf{P}_{w,j} - \sum_{k=1}^{q} \mathbf{P}_{l,k}, \\
    \label{equ:alpha_v}
    & \alpha_{\mathrm{D}} = \sigma(\delta) / \sigma(\bar{\delta}),
\end{align}
where $\alpha_{\mathrm{D}}$ denotes the data hardness, $\delta$ measures the difference between $\mathbf{P}_{w}$ and $\mathbf{P}_{l}$, and $\bar{\delta}$ denotes the mean difference across the dataset. The Sigmoid function $\sigma(\cdot)$ is employed to transform the response divergence $\mathbf{P}_{w}$ and the mean $\bar{\delta}$ into the range $(0, 1)$ for convenient comparison.

Finally, we adapt $\beta$ of Equ~\eqref{equ:dpo} to incorporate the hardness into the optimization procedure, and each preference instance corresponding to a specific $\beta$ as:
\begin{equation}
\label{equ:update beta data}
    \beta_{\mathrm{D}} = \beta \cdot \alpha_{\mathrm{D}}.
\end{equation}
This adjustment allows the model to optimize based on the data hardness, further enhancing its adaptability to the data.

\begin{figure}[t]
\vskip 0.2in
\begin{center}
\centerline{\includegraphics[width=\columnwidth]{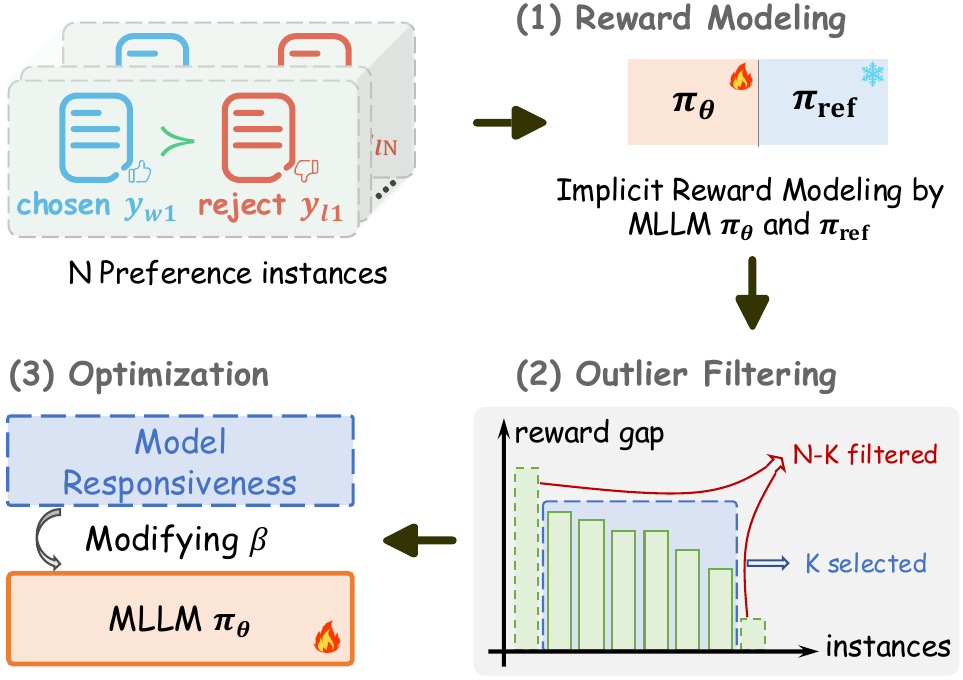}}
\caption{Overview of our model-aware preference optimization. Given $N$ preference instances: (1) we first calculate the reward gap of each instance using the implicit reward model; (2) To ensure stable modeling, we filter out the outliers (\textit{i.e.} the instance with excessively high or low gaps) and then estimate the average gap; (3) To enable the model to be aware of its current responsiveness, we integrate such estimation into the preference optimization process by modifying $\beta$ in Equ~\eqref{equ:dpo}.}
\label{fig:pipleine-llm-v2}
\end{center}
\vskip -0.2in
\end{figure}

\subsection{Model-aware Preference Optimization}
\label{subsec:Model-aware Preference Optimization}
An overview of our model-aware preference optimization is shown in Figure~\ref{fig:pipleine-llm-v2}. 
Given a batch of preference instances from dataset $\mathcal{D}$, as $\mathcal{B} = \{ (\mathcal{I}_{i}, x_{i}, y_{w,i}, y_{l,i}) | i = 1,2,\dots,N \} \sim \mathcal{D}$, 
it firstly calculates the reward gaps between the preferred $y_{w,i}$ and rejected $y_{l,i}$, and then filters out the outliers (\textit{i.e.} the instance with excessively high or low gaps) for stable estimation.
Such estimations are embedded into the preference optimization process by integrating into $\beta$ in Equ~\eqref{equ:dpo}, enabling the model to be aware of its current responsiveness. 
Details of model-aware preference optimization are as follows.

We employ current implicit reward gaps between the preferred and rejected responses of the given $\mathcal{B}$ instances to measure the current model's responsiveness. Specifically, the reward gap $\mathcal{R}$ for the $i$-th instance in $\mathcal{B}$ is formalized as:
\begin{equation}
\label{equ:hardness}
\resizebox{.91\hsize}{!}{
    \(\mathcal{R}_i =\left [  \beta \log \frac{{\pi}_{\bm{\theta}}(y_{w,i}|\mathcal{I}_i,x_i)}{{\pi}_{\text{ref}}(y_{w,i}|\mathcal{I}_i,x_i)}
    -\beta \log \frac{{\pi}_{\bm{\theta}}(y_{l,i}|\mathcal{I}_i,x_i)}{{\pi}_{\text{ref}}(y_{l,i}|\mathcal{I}_i,x_i)} \right ],\)
}
\end{equation}
where ${\pi}_{\bm{\theta}}$ and ${\pi}_{\text{ref}}$ represent the optimizing model and reference model, respectively. 
We then normalize the reward gaps using the estimated mean as follows:
\begin{equation}
\label{equ:batch_hard}
    \bar{{\mathcal{R}}_i} = \mathcal{R}_i / \bar{\mathcal{R}},
\end{equation}
where $\bar{\mathcal{R}}$ represents the estimated average reward gap, and $\bar{{\mathcal{R}}_i}$ is the normalized one for the $i$-th instance.

However, the estimation remains sensitive to outliers despite normalization, especially in the full fine-tuning scenario, where the batch size is relatively small.
To mitigate this issue, we filter out instances with exceptionally high or low gaps using a mask vector $\mathcal{M} \in \mathbb{R}^{N}$, defined as:
\begin{equation}
\label{equ:filter hardness}
    \bm{M}_{i} =
    \left\{
    \begin{aligned}
    &1, & (\bar{\mathcal{R}}_{i} - \bar{\mathcal{R}})^{2} \le  \tau, \\ 
    &0, & (\bar{\mathcal{R}}_{i} - \bar{\mathcal{R}})^{2}  >   \tau,
    \end{aligned}
    \right. 
\end{equation}
where $(\bar{\mathcal{R}_{i}} - \bar{\mathcal{R}})^{2}$ implies the squared distances from the mean, and $\tau$ represents the sorted $K$-th distance.
With the filtering, current responsiveness of ${\pi}_{\bm{\theta}}$ can be formalized as:
\begin{align}
\label{equ:average hardness}
    \bar{\mathcal{R}}_{\mathcal{B}} & = \frac{1}{N-K} \sum_{i=1}^{N}  \mathcal{M}_{i} \times \bar{\mathcal{R}}_{i}, \\
        \label{equ:alpha_M}
    \alpha_{\text{M}} & = \sigma(\bar{\mathcal{R}}_{\mathcal{B}}) / \sigma(\bar{\mathcal{R}}).
\end{align}

$\sigma(\cdot)$ is the Sigmoid function, which transforms both the filtered gaps $\bar{\mathcal{R}}_{\mathcal{B}}$ and the estimated mean $\bar{\mathcal{R}}$ into the range $(0, 1)$ for convenient comparison, and $\alpha_{M}$ refers to the estimated model responsiveness.

We then integrate $\alpha_{M}$ into the optimization procedure by modifying $\beta$ of Equ~\eqref{equ:dpo} as:
\begin{equation}
\label{equ:update beta reward}
    \beta_{\text{M}} = \beta \cdot \alpha_{\text{M}}.
\end{equation}
By utilizing $\beta_{M}$ to optimize $\pi_{\theta}$ with the filtered batch $\mathcal{B} \cdot \mathbf{M}$, the model can effectively adapt to its current responsiveness to the preference data.

Finally, we update the estimated mean $\bar{\mathcal{R}}$ using a moving average after optimization over batch $B$ as follows:
\begin{equation}
\label{equ:update mean}
    \bar{\mathcal{R}} \bm{\leftarrow} \gamma \cdot  \bar{\mathcal{R}} + (1- \gamma) \cdot \bar{\mathcal{R}}_{\mathcal{B}},
\end{equation}
the momentum $\gamma$ is set to $0.9$, and $\bar{\mathcal{H}}$ is initialized to $0$.

\subsection{Combining for Preference Optimization}
\label{subsec:combination}
In this section, we present the combination of both data- and model-aware strategies for robust preference optimization.
Specifically, given a batch of preference instances $\mathcal{B} = \{ (\mathcal{I}_{i}, x_{i}, y_{w, i}, y_{l, i}) | i = 1,2,\dots, N \} \sim \mathcal{D}$, where $\mathcal{I}_{i}$, $x_{i}$, $y_{w, i}$, $y_{l, i}$ denotes the image, question, preferred response, and rejected response, respectively,
we first compute data hardness offline based on our data-aware preference optimization, and then obtain the corresponding instance-wise hardness by Equ~\eqref{equ:alpha_v} as $\alpha_{\mathrm{D}}^{\mathcal{B}} \in \mathbb{R}^{N}$. 
Next, we calculate the reward gaps using Equ~\ref{equ:alpha_M} as $\alpha_{\text{M}}$.
To facilitate the optimization process to be both data- and model-aware, we propose an element-wise combination strategy. Specifically, we combine $\alpha_{\mathrm{D}}^{\mathcal{B}}$ and $\alpha_{\text{M}}$ as:
\begin{equation}
    \label{equ:alpha}
    \alpha = \alpha_{\mathrm{D}}^{\mathcal{B}} \cdot \alpha_{\text{M}},
\end{equation}
where $\alpha \in \mathbb{R}^{N}$ represents the combined factor. 
Subsequently, we adjust $\beta$ of Equ~\eqref{equ:dpo} to incorporate both components into optimization procedure as:
\begin{equation}
    \label{equ:update beta all}
        \beta_{\text{C}} = \beta \cdot \alpha,
    \end{equation}
where $\beta_{\text{C}} \in \mathbb{R}^{N}$, and each preference instance in $\mathcal{B}$ corresponds to a specific $\beta$. 
Finally, our combination strategy can be achieved by employing the $\beta_{\text{C}}$ to optimize $\pi_{\bm{\theta}}$ with the filtered batch $\mathcal{B} \cdot \bm{M}$, where $\bm{M}$ is obtained by Equ~\eqref{equ:filter hardness}.
Thus, the optimization process can become more adaptive, allowing the model to refine its preferences based on both pre-computed data hardness and real-time model responsiveness, further enhancing the robustness.

\begin{algorithm}[t]
   \caption{Algorithm of DAMA.}
   \label{algorithm}
\begin{algorithmic}
   \STATE {\bfseries Input:} Preference dataset $\mathcal{D}$, hyper-parameter $\beta$, SFT model $\pi_{\text{SFT}}$, CLIP classifier $\Gamma_{\text{CLIP}}$.
   \STATE {\bfseries Output}: The optimized model $\pi_{\bm{\theta}}$.
   \STATE Initialize model $\pi_{\bm{\theta}}$ and reference model $\pi_{\text{ref}}$ as $\pi_{\text{SFT}}$.
   \FOR{$\{ (\mathcal{I}, x, y_{w}, y_{l})\}$ in $\mathcal{D}$}
        \STATE 
        $\mathbf{S}_{w} \gets \text{LLM}\{y_w\}$, $\mathbf{S}_{l} \gets \text{LLM}\{y_l\}$;
        \STATE obtains $\delta$ with $\{\mathcal{I}, \mathbf{S}_{w}\}$, $\{\mathcal{I}, \mathbf{S}_{l}\}$; \COMMENT{Equ~\eqref{equ:subsentence-score} $\to$ \eqref{equ:difference}};
        \STATE $\alpha_{\mathrm{D}} \gets \sigma(\delta) / \sigma(\bar{\delta})$; \COMMENT{Equ~\eqref{equ:alpha_v}};
    \ENDFOR
   \REPEAT
   \FOR{$\mathcal{B} = \{ (\mathcal{I}_{i}, x_{i}, y_{w,i}, y_{l,i})\}_{i=1}^{N} \sim \mathcal{D}$}
        \STATE obtain $\mathcal{R}_{i}$ with $y_{w,i}$ and $y_{l,i}$; \COMMENT{Equ~\eqref{equ:hardness}};
        \STATE obtain $\bar{\mathcal{R}}_{\mathcal{B}}$ with $\mathcal{R}_{i}$;
        \COMMENT{Equ~\eqref{equ:batch_hard} $\to$ \eqref{equ:average hardness}};
        \STATE $\alpha_{M}  \gets \sigma(\bar{\mathcal{R}}_{\mathcal{B}}) / \sigma(\bar{\mathcal{R}})$; \COMMENT{Equ~\eqref{equ:alpha_M}};
        \STATE $\alpha \gets \alpha_{\mathrm{D}}^{\mathcal{B}} \cdot \alpha_{M}$, where $ \alpha_{\mathrm{D}}^{\mathcal{B}} = \{ {\alpha_{\mathrm{D,i}}}\}_{i=1}^{N}$;
        \COMMENT{Equ~\eqref{equ:alpha}}; 
        \STATE $\beta_{\mathrm{C}} \gets \beta \cdot \alpha$; \COMMENT{Equ~\eqref{equ:update beta all}}; 
        \STATE Compute loss {w.r.t.} $\beta_{\mathrm{C}}, \pi_{\bm{\theta}}$; \COMMENT{Equ~\eqref{equ:dpo}};
        \STATE Compute the gradient and update the model $\pi_{\bm{\theta}}$.
        \STATE $\bar{\mathcal{R}} \bm{\leftarrow} \gamma \cdot  \bar{\mathcal{R}} + (1- \gamma) \cdot \bar{\mathcal{R}}_{\mathcal{B}}$; \COMMENT{Equ~\eqref{equ:update mean}};
    \ENDFOR
   \UNTIL{The optimization is converged.}
\end{algorithmic}
\end{algorithm}

\section{Experiment}
\label{sec:experiment}
In this section, we elaborate on the effectiveness of our \textbf{Da}ta- and \textbf{M}odel-\textbf{a}ware Direct Preference Optimization (DAMA). Specifically, we first introduce the details of our experimental settings. Next, we illustrate the ablation studies, and finally, we compare the results with the state-of-the-art methods over various benchmarks.

\subsection{Experimental Settings}
In this section, we describe the experimental settings.
\newline
\noindent \textbf{Backbone}: We employ the LLaVA-1.5 7B and 13B for performance comparison \cite{LLaVA}.
\newline
\noindent \textbf{Dataset}: Our focus is not on the preference data construction, thus we directly utilize the released dataset by \cite{RLAIF-V}, which contains 22k preference data totally.

\noindent \textbf{Baselines.}
In this work, we compare against state-of-the-art baselines across various categories:
\newline
(1) {Hallucination-specific baselines.} In this category, we mainly compare with VCD \cite{VCD}, Less-is-more \cite{Less_Is_More}, OPERA \cite{OPERA}, and CCA-LLaVA \cite{CCA-LLaVA}.
\newline 
(2) {Preference Optimization-based baselines.} In this category, we mainly compare with HA-DPO \cite{HA-DPO}, POVID \cite{POVID}, LLaVA-RLHF \cite{LLaVA-RLHF}, RLHF-V \cite{RLHF-V}, RLAIF-V \cite{RLAIF-V}, AMP-MEG \cite{AMP-MEG}, CSR \cite{CSR}, V-DPO \cite{V-dpo}, and TPO \cite{TPO}.
\newline
(3) {Commercial baseline.} We include GPT-4V as a strong reference to evaluate the performance gap between the open-source and commercial models.

\noindent \textbf{Benchmarks.}
We conduct experiments on five benchmarks, including three hallucination benchmarks reflecting trustworthiness, and two general benchmarks:
\newline
(1) {Object HalBench} is employed to evaluate object hallucination by detailed descriptions of the image content. We report the response-level and mentioned-level \textbf{non}-hallucination rates to evaluate its capability to reduce hallucination \cite{object_hallucination_benchmark}.
\newline
(2) {AMBER} is a multi-dimensional hallucination benchmark, which contains more than 15k samples. We report the Accuracy and F1 metric by its discriminative component \cite{AMBER}.
\newline
(3) {MMHal-Bench} assesses response-level hallucination rate and informativeness by GPT-4 compare model outputs with human responses and object labels \cite{LLaVA-RLHF}.
\newline
(4) {LLaVA Bench} consists of 24 images and 60 questions including conversation, detailed description, and complex reasoning ability \cite{LLaVA}.
\newline
(5) {MM-Vet} is designed to evaluate six integrated competencies, including OCR, recognition, knowledge, language generation,  spatial awareness, and math \cite{MM-vet}.

\noindent \textbf{Implementation Details.}
For both LLaVA-1.5 7B and 13B models, we employ full parameter-tuning over the preference dataset with four epochs. Specifically, for reproducibility, we adopt the same hyperparameters as provided in the official LLaVA GitHub repository \footnote{https://github.com/haotian-liu/LLaVA}. The batch size $N$ is set to $16$, the selected size $K$ is set to $12$, and the penalty hyperparameter $\beta$ is set to $0.1$ by following \cite{DPO, RLAIF-V}. All experiments are conducted with four A100 80GB GPUs, and four epochs of fine-tuning cost seven hours for both backbones.

\subsection{Ablation Studies}
In this section, we evaluate the effects of different components of our DAMA. To this end, we utilize the LLaVA-1.5 7B model as the backbone. For clear illustration, we report both the response and the mentioned-level non-hallucination rate on the Object Hallucination benchmark, where the non-hallucination rate is defined as $100\% -$ hallucination rate.
The following are detailed illustrations.

\noindent \textbf{Influences of different components.}
The performance of various components of DAMA is reported in Table~\ref{tab:ab1-components}, where ``DPO'' refers to Direct Preference Optimization \cite{DPO}, ``MDPO'' represents our Model-aware Preference Optimization, ``D$^{2}$PO'' denotes Data-aware Preference Optimization, and ``DAMA'' corresponds to our combined strategy. The experimental results demonstrate: (1) All strategies significantly outperform the baseline method (DPO), especially our final ``DAMA'', achieving more than $10\%$ response level performance gains, highlighting the effectiveness of our method; (2) Compared with ``MDPO'', the performance gain of ``D$^{2}$PO'' is relatively modest, suggesting that the quality of the preference data is already high, which further validates the efficacy of the preference data construction strategy \cite{RLAIF-V}.

\begin{table}[t]
\caption{Experimental results of different components of DAMA.}
\label{tab:ab1-components}
\vskip 0.15in
\begin{center}
\begin{small}
\begin{sc}
\begin{tabular}{l | c c }
\toprule
\multirow{2}{*}{\textbf{Method}} & \multicolumn{2}{c}{\textbf{Object HalBench}} \\
& Response ($\uparrow$) & Mention ($\uparrow$) \\
\midrule
LLaVA-1.5-7B & 47.75 & 73.08 \\
+DPO & 78.29 & 89.48  \\
\midrule
+D$^{2}$PO & \addvalue{82.54}{4.25}{6.5} & \addvalue{90.64}{1.16}{6.5} \\
+MDPO & \addvalue{88.00}{9.71}{6.5} & \addvalue{93.74}{4.26}{6.5}\\
\rowcolor{lightgray} +\textbf{DAMA} & \addvalue{\textbf{90.87}}{12.58}{8} & \addvalue{\textbf{95.33}}{5.85}{6.5} \\
\bottomrule
\end{tabular}
\end{sc}
\end{small}
\end{center}
\vskip -0.1in
\end{table}

\noindent \textbf{Influences of probability transformation in data-aware preference optimization.} Table~\ref{tab:ab3-clip_prob} summarizes the performance of different response inconsistency estimation strategies, where  ``CLIP Scores'' denotes that we directly estimate $\delta$ based on $\mathbf{C}_{w}$ and $\mathbf{C}_{l}$, with $\delta = \sum_{j=1}^{p} \mathbf{C}_{w,j} / \sum_{k=1}^{q}\mathbf{C}_{l,k}$, and ``CLIP Probs'' represents our strategy, which transforms $\mathbf{C}_{w}$ and $\mathbf{C}_{l}$ into probabilities. The results indicate that: firstly, both strategies improve the performance, underscoring the effectiveness of integrating data inconsistency into the optimization process, which allows the model to better handle varying levels of data hardness, thereby improving overall robustness; furthermore, transforming $\mathbf{C}{w}$ and $\mathbf{C}_{l}$ into probabilities yields a larger performance gain, as probabilities smooth the estimation of $\delta$, mitigating noise from large gaps between $\mathbf{C}{w}$ and $\mathbf{C}_{l}$, and preventing the influences by the outliers.
\begin{table}[t]
\caption{Experimental results of different preference inconsistency construction strategies.}
\label{tab:ab3-clip_prob}
\vskip 0.15in
\begin{center}
\begin{small}
\begin{sc}
\begin{tabular}{l | c c }
\toprule
\multirow{2}{*}{\textbf{Method}} & \multicolumn{2}{c}{\textbf{Object HalBench}} \\

& Response ($\uparrow$) & Mention ($\uparrow$) \\
\midrule
DPO & 78.29 & 89.48  \\
\midrule
+CLIP Scores & \addvalue{80.65}{2.36}{6.5} & \addvalue{89.62}{0.14}{6.5} \\
\rowcolor{lightgray} \textbf{+CLIP Probs} & \addvalue{\textbf{82.54}}{4.25}{6.5} & \addvalue{\textbf{90.64}}{1.16}{6.5}\\
\bottomrule
\end{tabular}
\end{sc}
\end{small}
\end{center}
\vskip -0.1in
\end{table}

\noindent \textbf{Effects of the data filtering in model-aware preference optimization.} Table \ref{tab:ab2-data_filter} presents the performance of different data filtering strategies, where ``No Filter'' denotes that we directly utilize the mean gaps to estimate the model state without filtering, ``Bottom'' shows that we remove the $N-K$ samples with the largest distances in the batch, ``Top'' is filtering the $N-K$ samples with the smallest distances, ``Bottom \& Top'' refers to our filtering strategy, which filters both extremes based on the squared distances. Specifically, we can observe that: firstly, filtering solely from the bottom or top leads to performance degradation, indicating that such data introduces bias in estimating the model state. Moreover, filtering only the bottom samples results in significant performance drops due to overfitting on the top-ranked data, which misguides the estimation of model responsiveness to focus excessively on potentially less representative instances. Furthermore, filtering both bottom and top samples yields performance improvements, demonstrating the effectiveness of our proposed strategy, as it balances the influence of extreme data points.

\begin{table}[t]
\caption{Experimental results of different filtering strategies.}
\label{tab:ab2-data_filter}
\vskip 0.15in
\begin{center}
\begin{small}
\begin{sc}
\begin{tabular}{l | c c }
\toprule
\multirow{2}{*}{\textbf{Method}} & \multicolumn{2}{c}{\textbf{Object HalBench}} \\

& Response ($\uparrow$) & Mention ($\uparrow$) \\
\midrule
No Filter & 86.66 & 92.62 \\
\midrule
Bottom & \minusvalue{76.36}{10.30}{7.4} & \minusvalue{87.20}{5.42}{6} \\
Top & \minusvalue{86.34}{0.32}{6} & \minusvalue{92.48}{0.14}{6} \\
\rowcolor{lightgray} \textbf{Bottom \& Top} & \addvalue{\textbf{88.00}}{1.34}{6.5} & \addvalue{\textbf{93.74}}{1.12}{6.5} \\
\bottomrule
\end{tabular}
\end{sc}
\end{small}
\end{center}
\vskip -0.1in
\end{table}

\begin{table*}[t]
\caption{Performance comparisons with state-of-the-art methods on different benchmarks. 
We report non-hallucination rates in different levels including response level (Non-Rsp.) and mentioned-level (Non-Men.) for Object HalBench \cite{object_hallucination_benchmark}. Hall. refers to the Hallucination Rate for MMHal Bench \cite{LLaVA-RLHF}.
The best results of all methods are indicated in \textbf{bold}, and the second best results are \uline{underlined}.
The compared results are sourced from \cite{RLAIF-V, TPO}, and the reported results of LLaVA-1.5, DPO, and DAMA are evaluated using GPT-4-turbo-2024-04-09.}
\label{tab:comparison_SOTA}
\vskip 0.15in
\begin{center}
\begin{small}
\begin{sc}
\scalebox{0.86}{
\begin{tabular}{l | c | c c | c c | c c | c | c }
\toprule
\multirow{3}{*}{\textbf{Method}} & \multirow{3}{*}{\textbf{Size}} & \multicolumn{2}{c|}{\hspace{-1mm}\textbf{Object}} & \multicolumn{2}{c|}{\multirow{2}{*}{\hspace{-3mm}\textbf{AMBER}}} & \multicolumn{2}{c|}{\textbf{MMHal-}} & {\textbf{LLaVA}} & \multirow{3}{*}{\hspace{-2mm}\textbf{MM-Vet}($\uparrow$)} \\

& & \multicolumn{2}{c|}{\hspace{-1mm}\textbf{HalBench}} & & & \multicolumn{2}{c|}{\textbf{Bench}} & {\textbf{Bench}($\uparrow$)} & \\

\cline{3-8} & & Non-Rsp.($\uparrow$) & Non-Men.($\uparrow$) & Acc($\uparrow$) & F1($\uparrow$) & Scores($\uparrow$) & Hall.($\downarrow$) & & \\

\midrule
\multicolumn{10}{l}{\multirow{1}*{\textbf{Method (Hallucination-specific)}}}\\
\midrule
VCD~\conf{CVPR'24} & 7B & 51.2 & 75.7 & 71.8 & 74.9 & 2.12 & 54.2 & 61.6 & - \\
Less-is-more~\conf{ACL'24} & 7B & 59.7 & 82.2 & 72.4 & 75.8 & 2.33 & 50.0 & - & - \\
OPERA~\conf{CVPR'24} & 7B & 54.9 & 77.7 & 75.2 & 78.3 & 2.15 & 54.2 & 61.3 & - \\
CCA-LLaVA~\conf{NIPS'24} & 7B & 53.3 & 76.2 & 77.7 & 81.9 & 1.92 & 61.5 & 64.3 & - \\
\midrule
\multicolumn{10}{l}{\multirow{1}*{\textbf{Method (Preference optimization)}}}\\
\midrule
HA-DPO~\conf{arXiv'23} & 7B & 60.1 & 80.1 & 75.2 & 79.9 & 1.98 & 60.4 & - & - \\
POVID~\conf{arXiv'24} & 7B & 51.9 & 75.6 & 82.9 & 87.4 & 2.08 & 56.2 & 68.2 & 31.7 \\
RLHF-V~\conf{CVPR'24} & 7B & - & - & 74.8 & 78.5 & 2.02 & 60.4 & 68.0 & 32.3 \\
RLAIF-V~\conf{arXiv'24} & 7B & \uline{89.5} & \uline{94.8} & 76.8 & 84.5 & \uline{2.95} & \textbf{32.3} & - & - \\
CSR~\conf{NIPS'24} & 7B & - & - & 73.2 & 76.1 & 2.05 & 60.4 & 68.9 & 31.0 \\
V-DPO~\conf{EMNLP'24} & 7B & - & - & - & 81.6 & 2.16 & 56.0 & - & - \\
TPO~\conf{arXiv'24} & 7B & - & - & 79.3 & 85.0 & 2.47 & 51.0 & 70.2 & 33.0 \\
TPO~\conf{arXiv'24} & 13B & - & - & \uline{83.9} & \uline{88.0} & 2.72 & 45.8 & 72.8 & 36.2 \\
LLaVA-RLHF~\conf{ACL'24} & 13B & 61.9 & 81.1 & 79.7 & 83.9 & 2.02 & 62.5 & \textbf{95.6} & - \\
RLHF-V~\conf{CVPR'24} & 13B & 87.8 & 92.5 & 72.6 & 75.0 & 2.45 & 51.0 & \uline{76.7} & \textbf{38.5} \\
AMP-MEG~\conf{NIPS'24} & 13B & 68.3 & 79.4 & 79.5 & 84.6 & \textbf{3.08} & \uline{36.5} & - & - \\
\midrule
LLaVA-1.5 & 7B & 47.8 & 71.2 & 73.9 & 77.7 & 1.95 & 63.5 & 62.3 & 31.6 \\
\hspace{1mm} + DPO & 7B & 78.3 & 89.5  & 75.5 & 79.2 & 2.15 & 57.0 & 64.6 & 30.4\\
\hspace{1mm} + DAMA & 7B & \textbf{90.9} & \textbf{95.3}  & 83.3 & {87.0} & {2.76} & {41.0} & 68.0 & 32.8 \\
\rowcolor{lightgray} \hspace{1mm} Improvements (\%) & & \textcolor{blue}{\textbf{+16.1\%}} & \textcolor{blue}{\textbf{+6.5\%}}  & \textcolor{blue}{\textbf{+10.3\%}} & \textcolor{blue}{\textbf{+9.8\%}} & \textcolor{blue}{\textbf{+28.4\%}} & \textcolor{blue}{\textbf{+28.1\%}} & \textcolor{blue}{\textbf{+5.3\%}} & \textcolor{blue}{\textbf{+7.9\%}} \\
\midrule
LLaVA-1.5 & 13B & 50.0 & 76.4 & 71.2 & 73.0 & 2.36 & 56.0 & 66.1 & 36.1 \\
\hspace{1mm} + DPO & 13B & 84.5 & 92.4  & 78.5 & 83.5 & 2.59 & 48.0 & 74.0 & 35.4 \\
\hspace{1mm} + DAMA & 13B & 89.1 & 94.4  & \textbf{84.3} & \textbf{88.1} & {2.89} & 43.0 & 75.1 & \uline{36.4} \\
\rowcolor{lightgray} \hspace{1mm} Improvements (\%) & & \textcolor{blue}{\textbf{+5.1 \%}}  & \textcolor{blue}{\textbf{+2.1 \%}} & \textcolor{blue}{\textbf{+ 5.8\%}} & \textcolor{blue}{\textbf{+ 4.6\%}} & \textcolor{blue}{\textbf{+15.4\%}} & \textcolor{blue}{\textbf{+10.4\%}} & \textcolor{blue}{\textbf{+1.4 \%}} & \textcolor{blue}{\textbf{+2.8 \%}} \\
\midrule
\textbf{GPT-4V} & - & 86.4 & 92.7 & 83.4 & 87.4 & 3.49 & 28.1 & 98.0 & 67.7 \\
\bottomrule
\end{tabular}
}
\end{sc}
\end{small}
\end{center}
\vskip -0.1in
\end{table*}

\begin{figure}[t]
\vskip 0.2in
\begin{center}
\centerline{\includegraphics[width=\columnwidth]{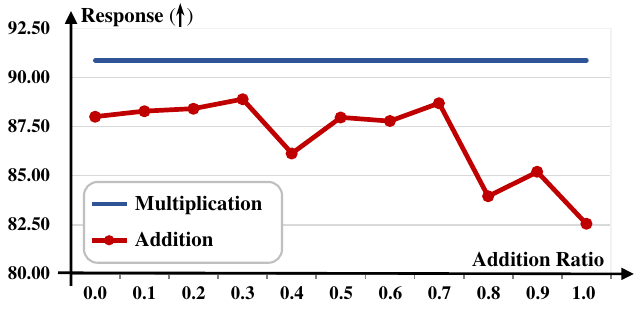}}
\caption{Experimental results of the combination strategies with the response-level non-hallucination rates.}
\label{fig:exp-ab-response}
\end{center}
\vskip -0.2in
\end{figure}

\begin{figure}[t]
\vskip 0.2in
\begin{center}
\centerline{\includegraphics[width=\columnwidth]{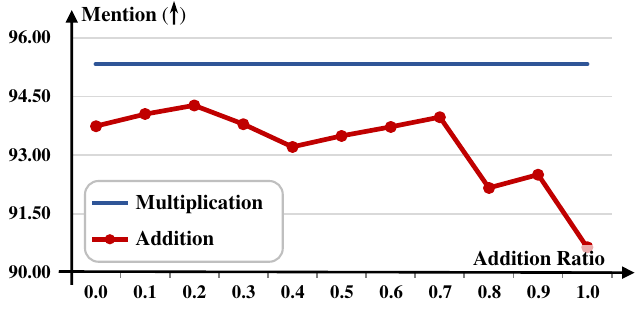}}
\caption{Experimental results of the combination strategies with the mentioned-level non-hallucination rates.}
\label{fig:exp-ab-mention}
\end{center}
\vskip -0.2in
\end{figure}

\noindent \textbf{Effects of the combination strategy.}
Figure~\ref{fig:exp-ab-response} and Figure ~\ref{fig:exp-ab-mention} illustrate the performance of different combination strategies from the Response and Mention, respectively. ``Multiplication'' refers to combining both effects using a weighted sum strategy, where $\alpha = (1-\rho) \cdot \alpha_{M} + \rho \cdot \alpha_{\mathrm{D}}^{\mathcal{B}}$, with $\rho$ ranging from $0.0$ to $1.0$, where $\rho = 0.0$ is the MDPO, and $\rho = 1.0$ refers to the ``D$^{2}$PO''. We can observe that multiplication outperforms the weighted sum by a significant margin. This is because the weighted sum balances the two components without effectively merging their influences, whereas multiplication amplifies their contributions, allowing for more robust and superior performance gains.

\subsection{Comparison with state-of-the-art methods}
In this subsection, we compare our method with state-of-the-art methods under three trustworthy benchmarks:  Object HalBench, AMBER, and MMHal-Bench, and two general benchmarks: LLaVA-Bench and MM-Vet. We compare our method against baselines from various aspects, including hallucination-specific baselines, preference optimization-based baselines, and the commercial baseline GPT-4V. Our DAMA is adapted to both the LLaVA-1.5 7B and 13B models. The experimental results, along with results of vanilla DPO and the improvements achieved by DAMA over DPO, are listed in Table \ref{tab:comparison_SOTA}. 

From the experimental results, we can observe that: (1) Compared to DPO, our DAMA achieves substantial performance improvements over all compared benchmarks. It's noteworthy that on MMHal-Bench, it achieves more than 10\% improvements for the 13B model, and more than 28\% gain for 7B models, which is significant; (2) DAMA achieves new state-of-the-art over various benchmarks. Our DAMA-7B reduces the response-level and mentioned-level hallucination on the generative Object HalBench by 90.9\% and 95.3\%. Moreover, DAMA-13B achieves 84.3\% Accuracy and 88.1\* F1 score on the discriminative AMBER benchmark. These attained results surpass those of GPT-4V, fully demonstrating the effectiveness of DAMA.
\section{Related Work}
\label{sec:related_work}

In this section, we overview the related background. Specifically, we first briefly summarize recent methods for aligning with human preferences, and then, we discuss existing hallucination mitigation methods for multi-modal large language models. Finally, we enumerate the differences between ours and related methods.

\subsection{Alignment with Human Preference}
To align the model outputs with human preferences, preference optimization methods have garnered significant attention \cite{rlhf, DPO, SimPO}. Concretely, RLHF (\cite{rlhf}) first introduces alignment through reward modeling, which trains a parametric reward model on preference data and subsequently optimizes the preference model using PPO \cite{PPO}. 
However, obtaining an effective reward model is challenging. To address this,  DPO \cite{DPO} is proposed to simplify the reward modeling process with an implicit reward function, allowing for direct optimization of preference data. Subsequent works like \cite{SimPO} and \cite{KTO} further simplify this process by removing the reference model and introducing binary feedback, respectively.

Current alignment works in MLLMs focus on two aspects:
(1) Collecting high-quality preference data. For example, existing works \cite{RLHF-V, RLAIF-V, VLFeedback} construct high-quality preference data to advance the research of MLLM alignment. For example, the works in \cite{RLHF-V, LLaVA-RLHF} introduce human-based preference construction strategies, RLAIF-V \cite{RLAIF-V} utilizes the open-source models (\textit{e.g.} LLaVA-NeXT \cite{LLaVA-Next}) to construct preference data, and the works in \cite{VLFeedback} and \cite{MAVIS} employ closed-source models like GPT-4V for preference annotation.
(2) Emphasizing the focus on visual detail. For instance, the works in \cite{V-dpo} and \cite{mDPO} construct rejected samples by destroying the image, the works like TPO \cite{TPO} and FiSAO \cite{FiSAO} identify and emphasize key tokens from the preference responses to attend to visual detail.

\subsection{Hallucination mitigation in MLLMs}
Hallucination, As a key indicator of trustworthiness, refers to that the MLLM outputs are not aligned with the image content \cite{hallucination-survey}. Beyond the preference optimization-based methods, current works to mitigate hallucination can be summarized into the following three aspects: 
(1) Data filtering. Works in LRV-Instruction \cite{LRV-Instruction} and Hallucidoctor \cite{Hallucidoctor} identify that noises within the instruction tuning data serve as a cause of hallucination, and introduce fine-grained data cleaning and curation strategies to mitigate hallucination.
(2) Enhanced visual representation. Works like \cite{Vcoder, MMVP} suggest that insufficient visual cues are the fundamental cause of hallucination, and they incorporate more intricate visual features to enrich the visual representations.
(3) Inference-time enhancement. Works like VCD \cite{VCD} and MARINE \cite{CFG} introduce visual contrastive decoding mechanisms to help enhance the model's focus on visual details by contrastively sampling from original and visually distorted distribution.

Compared with existing methods, which focus on data curation and introduce fine-grained regularizations, our proposed method aims to optimize adaptively based on the specific model responsiveness and the hardness of the data. 
We seek to design an improved optimization strategy that can effectively leverage the current model and the preference data, enabling the model to adaptively respond to the preference data, and thereby enhancing the alignment performance.
\section{Conclusion}
\label{sec:conclusion}

In this paper, to address the imbalanced responsiveness to data with varying hardness, we propose \textbf{Da}ta- and \textbf{M}odel-\textbf{a}ware Direct Preference Optimization (DAMA) to adaptively adjust the model's learning behavior from two aspects: (1) a data-aware strategy is proposed to dynamically adapt by incorporating the data hardness; and (2) a model-aware strategy is introduced to adaptively adjust by integrating the model's real-time responsiveness. Extensive experiments on five benchmarks from distinct aspects demonstrate the effectiveness of our method.
\newline
\noindent \textbf{Limitation and Future Work.} Our focus is on MLLMs handling images and text; future work should explore scalability to broader applications, including video and audio processing, as well as commercial multi-modal large language models, to fully demonstrate its practical impact.

\section*{Impact Statement}
\label{sec:impact}

This paper aims to enhance the robustness of multi-modal large language model alignment, thereby advancing the field of visual understanding. By tackling challenges like hallucination, we strive to improve the reliability of AI systems in alignment with human preferences.
We hope that our research will contribute to facilitating the development of AI systems that are both effective and reliable, ultimately delivering meaningful societal benefits.

\normalem
\balance{
\bibliography{main}

\begin{thebibliography}{39}
\providecommand{\natexlab}[1]{#1}
\providecommand{\url}[1]{\texttt{#1}}
\expandafter\ifx\csname urlstyle\endcsname\relax
  \providecommand{\doi}[1]{doi: #1}\else
  \providecommand{\doi}{doi: \begingroup \urlstyle{rm}\Url}\fi

\bibitem[Bai et~al.(2023)Bai, Bai, Yang, Wang, Tan, Wang, Lin, Zhou, and Zhou]{QwenVL}
Bai, J., Bai, S., Yang, S., Wang, S., Tan, S., Wang, P., Lin, J., Zhou, C., and Zhou, J.
\newblock Qwen-vl: A versatile vision-language model for understanding, localization, text reading, and beyond.
\newblock \emph{arXiv preprint arXiv:2308.12966}, 1\penalty0 (2):\penalty0 3, 2023.

\bibitem[Bradley \& Terry(1952)Bradley and Terry]{BT_model}
Bradley, R.~A. and Terry, M.~E.
\newblock Rank analysis of incomplete block designs: I. the method of paired comparisons.
\newblock \emph{Biometrika}, 39\penalty0 (3/4):\penalty0 324--345, 1952.

\bibitem[Chen et~al.(2024)Chen, Wu, Wang, Su, Chen, Xing, Zhong, Zhang, Zhu, Lu, et~al.]{InternVL}
Chen, Z., Wu, J., Wang, W., Su, W., Chen, G., Xing, S., Zhong, M., Zhang, Q., Zhu, X., Lu, L., et~al.
\newblock Internvl: Scaling up vision foundation models and aligning for generic visual-linguistic tasks.
\newblock In \emph{Proceedings of the IEEE/CVF Conference on Computer Vision and Pattern Recognition}, pp.\  24185--24198, 2024.

\bibitem[Cui et~al.(2024)Cui, Zhang, Zhou, Chen, Deng, Yao, and Chua]{FiSAO}
Cui, C., Zhang, A., Zhou, Y., Chen, Z., Deng, G., Yao, H., and Chua, T.-S.
\newblock Fine-grained verifiers: Preference modeling as next-token prediction in vision-language alignment.
\newblock \emph{arXiv preprint arXiv:2410.14148}, 2024.

\bibitem[Dubey et~al.(2024)Dubey, Jauhri, Pandey, Kadian, Al-Dahle, Letman, Mathur, Schelten, Yang, Fan, et~al.]{LLaMA3}
Dubey, A., Jauhri, A., Pandey, A., Kadian, A., Al-Dahle, A., Letman, A., Mathur, A., Schelten, A., Yang, A., Fan, A., et~al.
\newblock The llama 3 herd of models.
\newblock \emph{arXiv preprint arXiv:2407.21783}, 2024.

\bibitem[Ethayarajh et~al.(2024)Ethayarajh, Xu, Muennighoff, Jurafsky, and Kiela]{KTO}
Ethayarajh, K., Xu, W., Muennighoff, N., Jurafsky, D., and Kiela, D.
\newblock Kto: Model alignment as prospect theoretic optimization.
\newblock \emph{arXiv preprint arXiv:2402.01306}, 2024.

\bibitem[Gu et~al.(2024)Gu, Wang, Cao, Bu, Song, He, Li, and Zheng]{TPO}
Gu, J., Wang, Y., Cao, M., Bu, P., Song, J., He, Y., Li, S., and Zheng, B.
\newblock Token preference optimization with self-calibrated visual-anchored rewards for hallucination mitigation.
\newblock \emph{arXiv preprint arXiv:2412.14487}, 2024.

\bibitem[Huang et~al.(2024)Huang, Dong, Zhang, Wang, He, Wang, Lin, Zhang, and Yu]{OPERA}
Huang, Q., Dong, X., Zhang, P., Wang, B., He, C., Wang, J., Lin, D., Zhang, W., and Yu, N.
\newblock Opera: Alleviating hallucination in multi-modal large language models via over-trust penalty and retrospection-allocation.
\newblock In \emph{Proceedings of the IEEE/CVF Conference on Computer Vision and Pattern Recognition}, pp.\  13418--13427, 2024.

\bibitem[Jain et~al.(2024)Jain, Yang, and Shi]{Vcoder}
Jain, J., Yang, J., and Shi, H.
\newblock Vcoder: Versatile vision encoders for multimodal large language models.
\newblock In \emph{Proceedings of the IEEE/CVF Conference on Computer Vision and Pattern Recognition}, pp.\  27992--28002, 2024.

\bibitem[Leng et~al.(2024)Leng, Zhang, Chen, Li, Lu, Miao, and Bing]{VCD}
Leng, S., Zhang, H., Chen, G., Li, X., Lu, S., Miao, C., and Bing, L.
\newblock Mitigating object hallucinations in large vision-language models through visual contrastive decoding.
\newblock In \emph{Proceedings of the IEEE/CVF Conference on Computer Vision and Pattern Recognition}, pp.\  13872--13882, 2024.

\bibitem[Li et~al.(2024)Li, Xie, Li, Chen, Wang, Chen, Yang, Wang, Kong, and Liu]{VLFeedback}
Li, L., Xie, Z., Li, M., Chen, S., Wang, P., Chen, L., Yang, Y., Wang, B., Kong, L., and Liu, Q.
\newblock Vlfeedback: A large-scale ai feedback dataset for large vision-language models alignment.
\newblock \emph{arXiv preprint arXiv:2410.09421}, 2024.

\bibitem[Liu et~al.(2023)Liu, Lin, Li, Wang, Yacoob, and Wang]{LRV-Instruction}
Liu, F., Lin, K., Li, L., Wang, J., Yacoob, Y., and Wang, L.
\newblock Mitigating hallucination in large multi-modal models via robust instruction tuning.
\newblock In \emph{The Twelfth International Conference on Learning Representations}, 2023.

\bibitem[Liu et~al.(2024{\natexlab{a}})Liu, Li, Li, Li, Zhang, Shen, and Lee]{LLaVA-Next}
Liu, H., Li, C., Li, Y., Li, B., Zhang, Y., Shen, S., and Lee, Y.~J.
\newblock Llava-next: Improved reasoning, ocr, and world knowledge, 2024{\natexlab{a}}.

\bibitem[Liu et~al.(2024{\natexlab{b}})Liu, Li, Wu, and Lee]{LLaVA}
Liu, H., Li, C., Wu, Q., and Lee, Y.~J.
\newblock Visual instruction tuning.
\newblock \emph{Advances in neural information processing systems}, 36, 2024{\natexlab{b}}.

\bibitem[Liu et~al.(2024{\natexlab{c}})Liu, Xue, Chen, Chen, Zhao, Wang, Hou, Li, and Peng]{hallucination-survey}
Liu, H., Xue, W., Chen, Y., Chen, D., Zhao, X., Wang, K., Hou, L., Li, R., and Peng, W.
\newblock A survey on hallucination in large vision-language models.
\newblock \emph{arXiv preprint arXiv:2402.00253}, 2024{\natexlab{c}}.

\bibitem[Meng et~al.(2024)Meng, Xia, and Chen]{SimPO}
Meng, Y., Xia, M., and Chen, D.
\newblock Simpo: Simple preference optimization with a reference-free reward.
\newblock \emph{arXiv preprint arXiv:2405.14734}, 2024.

\bibitem[Ouali et~al.(2025)Ouali, Bulat, Martinez, and Tzimiropoulos]{CLIP-DPO}
Ouali, Y., Bulat, A., Martinez, B., and Tzimiropoulos, G.
\newblock Clip-dpo: Vision-language models as a source of preference for fixing hallucinations in lvlms.
\newblock In \emph{European Conference on Computer Vision}, pp.\  395--413. Springer, 2025.

\bibitem[Ouyang et~al.(2022)Ouyang, Wu, Jiang, Almeida, Wainwright, Mishkin, Zhang, Agarwal, Slama, Ray, et~al.]{rlhf}
Ouyang, L., Wu, J., Jiang, X., Almeida, D., Wainwright, C., Mishkin, P., Zhang, C., Agarwal, S., Slama, K., Ray, A., et~al.
\newblock Training language models to follow instructions with human feedback.
\newblock \emph{Advances in neural information processing systems}, 35:\penalty0 27730--27744, 2022.

\bibitem[Radford et~al.(2021)Radford, Kim, Hallacy, Ramesh, Goh, Agarwal, Sastry, Askell, Mishkin, Clark, et~al.]{CLIP}
Radford, A., Kim, J.~W., Hallacy, C., Ramesh, A., Goh, G., Agarwal, S., Sastry, G., Askell, A., Mishkin, P., Clark, J., et~al.
\newblock Learning transferable visual models from natural language supervision.
\newblock In \emph{International conference on machine learning}, pp.\  8748--8763. PMLR, 2021.

\bibitem[Rafailov et~al.(2024)Rafailov, Sharma, Mitchell, Manning, Ermon, and Finn]{DPO}
Rafailov, R., Sharma, A., Mitchell, E., Manning, C.~D., Ermon, S., and Finn, C.
\newblock Direct preference optimization: Your language model is secretly a reward model.
\newblock \emph{Advances in Neural Information Processing Systems}, 36, 2024.

\bibitem[Rohrbach et~al.(2018)Rohrbach, Hendricks, Burns, Darrell, and Saenko]{object_hallucination_benchmark}
Rohrbach, A., Hendricks, L.~A., Burns, K., Darrell, T., and Saenko, K.
\newblock Object hallucination in image captioning.
\newblock \emph{arXiv preprint arXiv:1809.02156}, 2018.

\bibitem[Schulman et~al.(2017)Schulman, Wolski, Dhariwal, Radford, and Klimov]{PPO}
Schulman, J., Wolski, F., Dhariwal, P., Radford, A., and Klimov, O.
\newblock Proximal policy optimization algorithms.
\newblock \emph{arXiv preprint arXiv:1707.06347}, 2017.

\bibitem[Sun et~al.(2024)Sun, Shen, Cao, Liu, Li, Shen, Gan, Gui, Wang, Yang, et~al.]{LLaVA-RLHF}
Sun, Z., Shen, S., Cao, S., Liu, H., Li, C., Shen, Y., Gan, C., Gui, L.-Y., Wang, Y.-X., Yang, Y., et~al.
\newblock Aligning large multimodal models with factually augmented rlhf.
\newblock In \emph{Findings of the Association for Computational Linguistics}, 2024.

\bibitem[Tong et~al.(2024)Tong, Liu, Zhai, Ma, LeCun, and Xie]{MMVP}
Tong, S., Liu, Z., Zhai, Y., Ma, Y., LeCun, Y., and Xie, S.
\newblock Eyes wide shut? exploring the visual shortcomings of multimodal llms.
\newblock In \emph{Proceedings of the IEEE/CVF Conference on Computer Vision and Pattern Recognition}, pp.\  9568--9578, 2024.

\bibitem[Wang et~al.(2024)Wang, Zhou, Huang, Xu, Zhang, Poon, and Chen]{mDPO}
Wang, F., Zhou, W., Huang, J.~Y., Xu, N., Zhang, S., Poon, H., and Chen, M.
\newblock mdpo: Conditional preference optimization for multimodal large language models.
\newblock \emph{arXiv preprint arXiv:2406.11839}, 2024.

\bibitem[Wang et~al.(2023)Wang, Wang, Xu, Zhang, Gu, Jia, Yan, Zhang, and Sang]{AMBER}
Wang, J., Wang, Y., Xu, G., Zhang, J., Gu, Y., Jia, H., Yan, M., Zhang, J., and Sang, J.
\newblock An llm-free multi-dimensional benchmark for mllms hallucination evaluation.
\newblock \emph{arXiv preprint arXiv:2311.07397}, 2023.

\bibitem[Xie et~al.(2024)Xie, Li, Xu, and Kan]{V-dpo}
Xie, Y., Li, G., Xu, X., and Kan, M.-Y.
\newblock V-dpo: Mitigating hallucination in large vision language models via vision-guided direct preference optimization.
\newblock In \emph{Findings of the Association for Computational Linguistics: EMNLP}, pp.\  13258--13273, 2024.

\bibitem[Xing et~al.(2024)Xing, Li, Laptev, and Lu]{CCA-LLaVA}
Xing, Y., Li, Y., Laptev, I., and Lu, S.
\newblock Mitigating object hallucination via concentric causal attention.
\newblock \emph{Advances in neural information processing systems}, 2024.

\bibitem[Yu et~al.(2024{\natexlab{a}})Yu, Li, Wei, Pang, Ye, Qin, Tang, Tian, and Zhuang]{Hallucidoctor}
Yu, Q., Li, J., Wei, L., Pang, L., Ye, W., Qin, B., Tang, S., Tian, Q., and Zhuang, Y.
\newblock Hallucidoctor: Mitigating hallucinatory toxicity in visual instruction data.
\newblock In \emph{Proceedings of the IEEE/CVF Conference on Computer Vision and Pattern Recognition}, pp.\  12944--12953, 2024{\natexlab{a}}.

\bibitem[Yu et~al.(2024{\natexlab{b}})Yu, Yao, Zhang, He, Han, Cui, Hu, Liu, Zheng, Sun, et~al.]{RLHF-V}
Yu, T., Yao, Y., Zhang, H., He, T., Han, Y., Cui, G., Hu, J., Liu, Z., Zheng, H.-T., Sun, M., et~al.
\newblock Rlhf-v: Towards trustworthy mllms via behavior alignment from fine-grained correctional human feedback.
\newblock In \emph{Proceedings of the IEEE/CVF Conference on Computer Vision and Pattern Recognition}, pp.\  13807--13816, 2024{\natexlab{b}}.

\bibitem[Yu et~al.(2024{\natexlab{c}})Yu, Zhang, Yao, Dang, Chen, Lu, Cui, He, Liu, Chua, et~al.]{RLAIF-V}
Yu, T., Zhang, H., Yao, Y., Dang, Y., Chen, D., Lu, X., Cui, G., He, T., Liu, Z., Chua, T.-S., et~al.
\newblock Rlaif-v: Aligning mllms through open-source ai feedback for super gpt-4v trustworthiness.
\newblock \emph{arXiv preprint arXiv:2405.17220}, 2024{\natexlab{c}}.

\bibitem[Yu et~al.(2023)Yu, Yang, Li, Wang, Lin, Liu, Wang, and Wang]{MM-vet}
Yu, W., Yang, Z., Li, L., Wang, J., Lin, K., Liu, Z., Wang, X., and Wang, L.
\newblock Mm-vet: Evaluating large multimodal models for integrated capabilities.
\newblock \emph{arXiv preprint arXiv:2308.02490}, 2023.

\bibitem[Yue et~al.(2024)Yue, Zhang, and Jin]{Less_Is_More}
Yue, Z., Zhang, L., and Jin, Q.
\newblock Less is more: Mitigating multimodal hallucination from an eos decision perspective.
\newblock In \emph{Proceedings of the 62nd Annual Meeting of the Association for Computational Linguistics}, pp.\  11766--11781, 2024.

\bibitem[Zhang et~al.(2024{\natexlab{a}})Zhang, Wu, Lu, Song, Rong, Yao, Zhao, Liu, Sun, Feng, et~al.]{AMP-MEG}
Zhang, M., Wu, W., Lu, Y., Song, Y., Rong, K., Yao, H., Zhao, J., Liu, F., Sun, Y., Feng, H., et~al.
\newblock Automated multi-level preference for mllms.
\newblock \emph{Advances in Neural Information Processing Systems}, 2024{\natexlab{a}}.

\bibitem[Zhang et~al.(2024{\natexlab{b}})Zhang, Wei, Jiang, Guo, Li, Zhang, Tong, Liu, Zhou, Wei, et~al.]{MAVIS}
Zhang, R., Wei, X., Jiang, D., Guo, Z., Li, S., Zhang, Y., Tong, C., Liu, J., Zhou, A., Wei, B., et~al.
\newblock Mavis: Mathematical visual instruction tuning with an automatic data engine.
\newblock \emph{arXiv preprint arXiv:2407.08739}, 2024{\natexlab{b}}.

\bibitem[Zhao et~al.(2024)Zhao, Deng, Zhang, and Gu]{CFG}
Zhao, L., Deng, Y., Zhang, W., and Gu, Q.
\newblock Mitigating object hallucination in large vision-language models via classifier-free guidance.
\newblock \emph{arXiv preprint arXiv:2402.08680}, 2024.

\bibitem[Zhao et~al.(2023)Zhao, Wang, Ouyang, Dong, Wang, and He]{HA-DPO}
Zhao, Z., Wang, B., Ouyang, L., Dong, X., Wang, J., and He, C.
\newblock Beyond hallucinations: Enhancing lvlms through hallucination-aware direct preference optimization.
\newblock \emph{arXiv preprint arXiv:2311.16839}, 2023.

\bibitem[Zhou et~al.(2024{\natexlab{a}})Zhou, Cui, Rafailov, Finn, and Yao]{POVID}
Zhou, Y., Cui, C., Rafailov, R., Finn, C., and Yao, H.
\newblock Aligning modalities in vision large language models via preference fine-tuning.
\newblock \emph{arXiv preprint arXiv:2402.11411}, 2024{\natexlab{a}}.

\bibitem[Zhou et~al.(2024{\natexlab{b}})Zhou, Fan, Cheng, Yang, Chen, Cui, Wang, Li, Zhang, and Yao]{CSR}
Zhou, Y., Fan, Z., Cheng, D., Yang, S., Chen, Z., Cui, C., Wang, X., Li, Y., Zhang, L., and Yao, H.
\newblock Calibrated self-rewarding vision language models.
\newblock \emph{Advances in Neural Information Processing Systems}, 2024{\natexlab{b}}.

\end{thebibliography}
\bibliographystyle{icml2025}
}
\newpage
\appendix
\onecolumn
\section{Appendix}
\subsection{Qualitative analysis}
\begin{figure}[H]
\vskip 0.2in
\begin{center}
\centerline{\includegraphics[width=0.65\columnwidth]{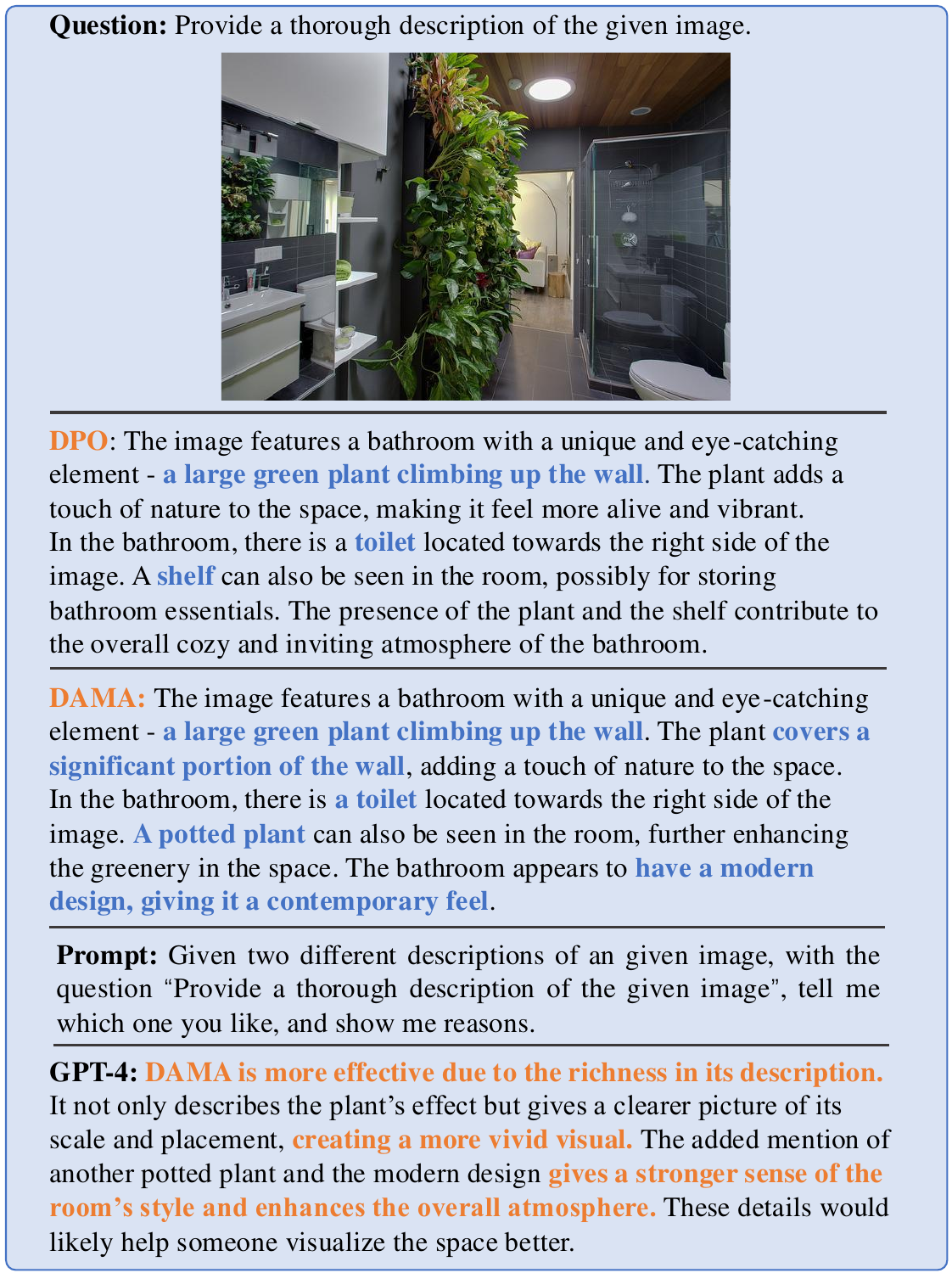}}
\caption{Qualitative Comparisons between DAMA and vanilla DPO. Data source: Object HalBench.}
\label{fig:demo-1}
\end{center}
\vskip -0.2in
\end{figure}

\begin{figure*}[t]
\vskip 0.2in
\begin{center}
\centerline{\includegraphics[width=0.65\columnwidth]{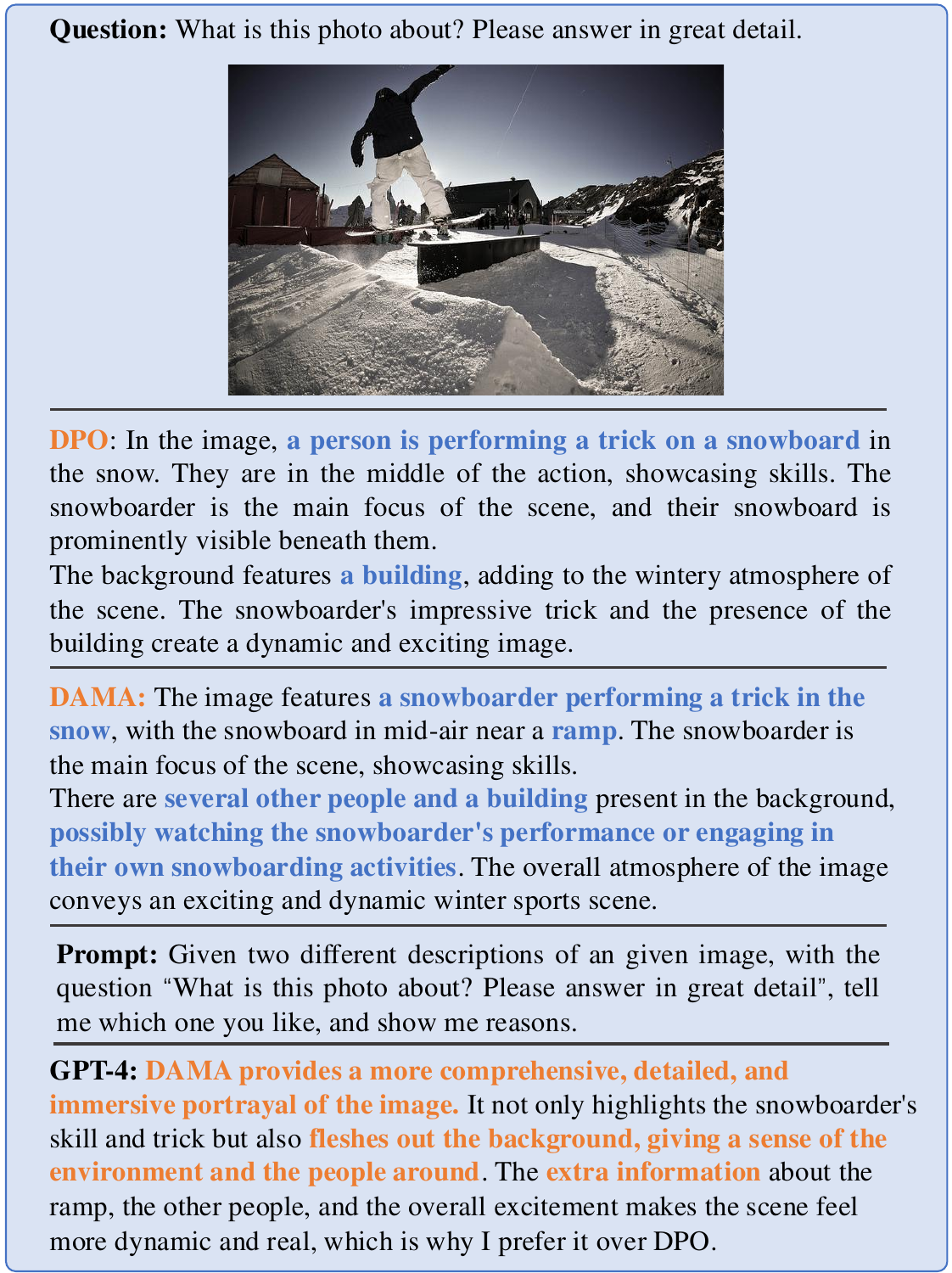}}
\caption{Qualitative Comparisons between DAMA and vanilla DPO. Data source: Object HalBench.}
\label{fig:demo-2}
\end{center}
\vskip -0.2in
\end{figure*}

In this section, we provide qualitative analysis between our Data- and Model-aware preference optimization (DAMA) and the DPO method \cite{DPO}. The case studies are shown in Figure \ref{fig:demo-1} and Figure \ref{fig:demo-2}, and we also include the evaluations of GPT-4. From the case studies, we can observe two key points: (1) while both methods generate correct outputs, DPO tends to be more general, whereas DAMA offers richer, more detailed descriptions. This can be attributed to the integration of data hardness, quantified by the CLIP score between the Image and responses, into the optimization process, enabling DAMA to capture and describe more nuanced visual details;
and (2) as evidenced by the GPT-4 evaluation, DAMA generates more vivid and immersive visual descriptions, this can be attributed to the enhancement of the responsiveness to the visual content, resulting in a more expressive and contextually aligned representation of the scene.

\end{document}